\documentclass[sigconf,authorversion,nonacm]{acmart}
\usepackage{pdfpages}
\usepackage{CJKutf8}
\usepackage{multirow}
\usepackage{makecell}
\usepackage{arydshln}
\usepackage{color}
\usepackage{diagbox}
\usepackage{booktabs}
\usepackage{enumitem}
\usepackage{bm}
\usepackage{marvosym}

\usepackage{bbding}

\theoremstyle{definition}

\theoremstyle{theorem}

\theoremstyle{proof}

\theoremstyle{remark}
\newtheorem*{remark}{Remark}

\AtBeginDocument{%
  \providecommand\BibTeX{{%
    \normalfont B\kern-0.5em{\scshape i\kern-0.25em b}\kern-0.8em\TeX}}}

\begin{document}

\title{Multi-Modal Experience Inspired AI Creation}

\author{Qian Cao$^{\dagger}$}
\email{caoqian4real@ruc.edu.cn}
\orcid{0000-0003-3288-1714}
\thanks{$\dagger$ Beijing Key Laboratory of Big Data Management and Analysis Methods.}
\affiliation{%
  \institution{Gaoling School of Artificial Intelligence, Renmin University of China}
  \country{ }
}

\author{Xu Chen$^{\dagger}$}
\email{successcx@gmail.com}
\affiliation{%
  \institution{Gaoling School of Artificial Intelligence, Renmin University of China}
  \country{ }
}

\author{Ruihua Song$^{\dagger}$\textsuperscript{\Letter}}
\email{rsong@ruc.edu.cn}
\thanks{\Letter\ Corresponding author.}
\affiliation{%
  \institution{Gaoling School of Artificial Intelligence, Renmin University of China}
  \country{ }
}

\author{Hao Jiang}
\email{jianghao66@huawei.com}
\affiliation{%
  \institution{Poisson Lab, Huawei}
  \country{ }
}

\author{Guang Yang}
\email{yangguang97@huawei.com}
\affiliation{%
  \institution{Poisson Lab, Huawei}
  \country{ }
}

\author{Zhao Cao}
\email{caozhao1@huawei.com}
\affiliation{%
  \institution{Poisson Lab, Huawei}
  \country{ }
}

\renewcommand{\authors}{Qian Cao, Xu Chen, Ruihua Song, Hao Jiang, Guang Yang, Zhao Cao}
\renewcommand{\shortauthors}{Qian Cao et al.}

\begin{abstract}
AI creation, such as poem or lyrics generation, has attracted increasing attention from both industry and academic communities, with many promising models proposed in the past few years.
Existing methods usually estimate the outputs based on single and independent visual or textual information.
However, in reality, humans usually make creations according to their experiences, which may involve different modalities and be sequentially correlated.
To model such human capabilities, in this paper, we define and solve a novel AI creation problem based on human experiences.
More specifically, we study how to generate texts based on sequential multi-modal information.
Compared with the previous works, this task is much more difficult because the designed model has to well understand and adapt the semantics among different modalities and effectively convert them into the output in a sequential manner.
To alleviate these difficulties, we firstly design a multi-channel sequence-to-sequence architecture equipped with a multi-modal attention network.
For more effective optimization, we then propose a curriculum negative sampling strategy tailored for the sequential inputs.
To benchmark this problem and demonstrate the effectiveness of our model, we manually labeled a new multi-modal experience dataset.
With this dataset, we conduct extensive experiments by comparing our model with a series of representative baselines, where we can demonstrate significant improvements in our model based on both automatic and human-centered metrics. The code and data are available
at: \url{https://github.com/Aman-4-Real/MMTG}.
\end{abstract}


\maketitle

\section{Introduction}\label{sec:intro}

AI creation, such as poem writing or lyrics generation, explores human high-level intelligence on languages, which is becoming an important research direction~\cite{yi2018chinese,liu2018beyond,malmi2016dopelearning} and has been successfully applied to many real-world applications~\cite{zhipeng2019jiuge,zhang2020youling}.
Usually, AI creation can be formalized as text generation tasks, where the input is either visual or textual information, and the output is a sequence of texts.
For example, to write vivid poems based on a given image,~\citet{liu2018beyond} proposes an adversarial reinforcement learning method to bridge the visual and textual spaces.
\citet{wang2016chinese} generates the texts based on many given topics line by line.

\begin{figure}[t]
    \centering
    \includegraphics[width=1.\linewidth]{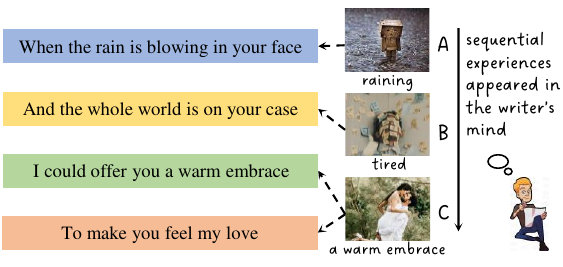}
    \caption{A toy example of the human creation process. The inputs and outputs are sequentially corresponded in a loose manner, that is, each input may influence multiple outputs.}
    \label{fig:intro}
\end{figure}

While the above models have achieved many successes, there are still many gaps between human and machine creation processes.
To begin with, humans usually perceive and understand the world through multi-channel information, such as seeing, listening, or touching the objects around them.
However, most existing AI creation models still fall into the single modality generation paradigm, either from images to texts or from texts to texts.
In addition, humans usually make creations according to their dynamic and ``sequential experiences''. 
Here ``experience'' indicates the feelings stored in the writers' minds and ``sequential'' means the past ``experiences'' are evoked in chronological or logical order in the writers' creation process.
For example, writers may describe rain blowing (a kind of visual experience) when they memorize a tough day, and accompanying the feeling of being tired (a kind of textual experience) in some order (see Figure~\ref{fig:intro}). 
Previously, the AI creation models usually output text sequences based on a single input\cite{liu2018beyond}, or do not consider the orders among the multiple inputs\cite{liu2018images2poem}, which fail to capture human capabilities in controlling and preserving the sequential information when making creations.

To fill up the above gaps, in this paper, we define a novel AI creation task that, given a topic and a sequence of image and text pairs which simulate the multi-modal experience in mind, the goal is to generate a text sequence describing the input multi-modal information and simultaneously preserve the sequential semantics of the inputs.
Compared with the previous AI creation problems, this task makes a further step towards more realistic human creation processes.
However, at the same time, it is much more challenging because:
(1) unlike previous work, where the input is a single modality, in our task, there is topical, visual, and textual information as not all experiences correspond to visual concepts. How to combine them and adaptively convert them into a text sequence needs our dedicated designs.
(2) In our problem, the sequential correspondences between the inputs and outputs are of great importance.
However, unlike common sequence to sequence problems~{\cite{sutskever2014sequence,yi2017generating}}, where there are rigorous input-output correspondences, human creations are much more flexible, and each input may influence multiple outputs. As exampled in Figure~\ref{fig:intro}, image C, which plots the embracing between a couple of lovers, determines the third and fourth output sentences.
Thus, how to properly design models to encode the above heuristics is another challenge.
(3) To the best of our knowledge, there is no publicly available dataset for our sequential multi-modal AI creation task, which makes it hard to evaluate whether the proposed solutions are effective.

To solve the above challenges, we design a multi-channel sequence-to-sequence architecture, where different modalities are firstly projected into the same space based on the attention mechanism, and then the output is recurrently generated based on the fused information.
In order to control the influence of the input on the output, we design a novel self-attention method to enable the incorporation of heuristic prior knowledge.
In addition, to more effectively optimize the sequence-to-sequence model, we propose a curriculum learning based negative sampling strategy to schedule the training samples in an ``easy-to-hard'' manner. 
To verify the superiority of the proposed model, we manually label a new dataset for the task of sequential multi-modal AI creation.
Empirical studies demonstrate that our model can be more effective as compared with the baselines in terms of both automatic and human evaluation metrics.

The main contributions of this paper can be concluded as follows:

$\bullet$ Inspired by the real human creation process, we formulate a novel AI creation task, where the output text should be generated based on the multi-modal input information and consider the sequential semantics of the inputs.

$\bullet$ To solve the above problem, we design a neural sequence-to-sequence model and also enhance it with a prior knowledge guided self-attention module and a curriculum negative sampling strategy.

$\bullet$ We build the first dataset for the above task, and conduct extensive experiments to verify the effectiveness of our proposed method by comparing it with the state-of-the-art baselines.

\begin{figure*}[t]
    \centering
    \includegraphics[width=.945\linewidth]{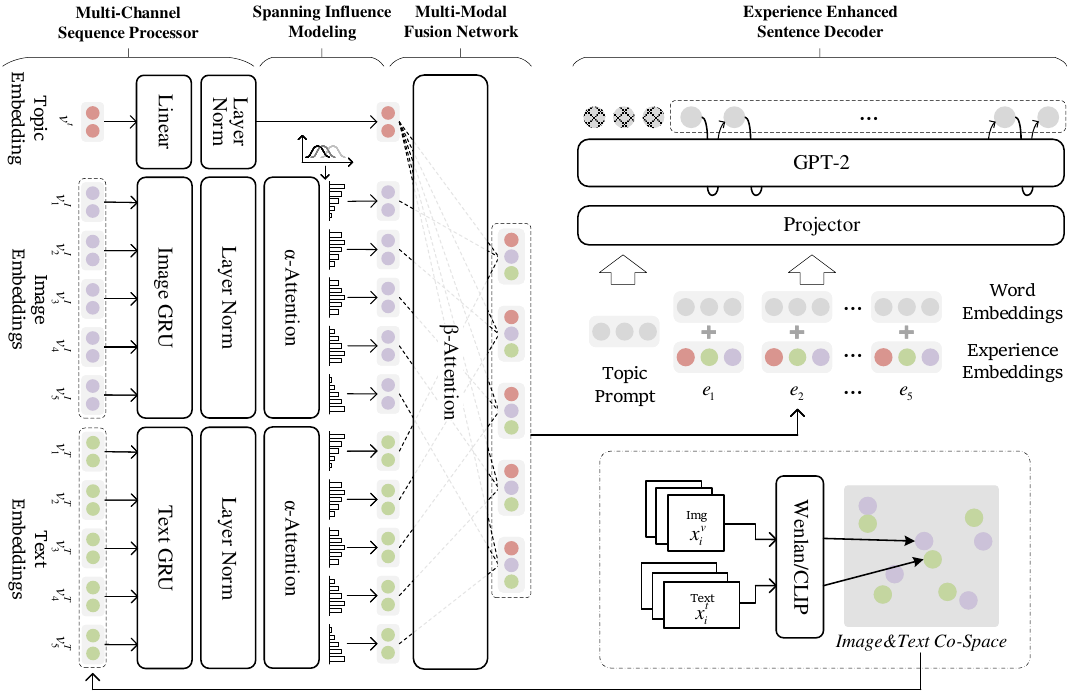}
    \caption{The framework of our proposed MMTG model. Experiences are shown in image and text sequences. An image corresponds to its text at the same time step. The modules of Multi-Channel Sequence Processor, Spanning Influence Modeling, Multi-Modal Fusion Network, and Experience Enhanced Sentence Decoder are presented from left to right.}
    \label{fig:structure}
\end{figure*}

\section{Problem Formulation}\label{formu}
Basically, our problem aims to predict a sequence of sentences given a topic and a set of ordered image-text pairs.
Formally, for each sample, suppose the input topic and set of image-text pairs are $\bm{t}_i$ and $\{(\bm{x}_{i,j}^I, \bm{x}_{i,j}^T)\}_{j=1}^{L}$, respectively, where $\bm{x}_{i,j}^I$ and $\bm{x}_{i,j}^T$ are the image and text at each step, and $L$ is the length of the set. In our paper, we construct the set of image-text pairs, and we can also retrieve image-text pairs by topics in real applications. 
The text $\bm{x}_{i,j}^T$ is composed of a series of words $\{{x}_{i,j,1}^T, {x}_{i,j,2}^T,...,{x}_{i,j,l_{ij}}^T\}$, and $l_{ij}$ is the length of the text.
Let $\bm{y}_i = \{\bm{y}_{i,k}\}_{k=1}^L$ be the output sentence set, where $\bm{y}_{i,k} = \{{y}_{i,k,1}, {y}_{i,k,2},...,{y}_{i,k,d_{ik}}\}$ is the word sequence for the $k$-th sentence, and $d_{ik}$ is the length of the sequence.
We denote by $\mathcal{S} = \{(\bm{t}_i$, $\{(\bm{x}_{i,j}^I, \bm{x}_{i,j}^T)\}_{j=1}^{L}), \bm{y}_i\}_{i=1}^N$ the training set. Then given $\mathcal{S}$, we aim to learn a model $f$, which can accurately predict a sequence of sentences given a topic and a set of image-text pairs in the testing set.
In the following sections, for simplicity, we omit the sample index $i$ when there is no confusion.

\section{Methodology}
In this section, we detail our framework, which is illustrated in Figure~\ref{fig:structure}.
In general, our framework is composed of four parts.
The first three parts form the encoder.
In specific, the raw images and texts are firstly handled by a multi-channel sequence processor to produce their semantic embeddings.
Then, the embedding at each step is separated into different parts to influence the final output.
At last, different modalities are fused with an attention network.
The last part is the decoder, aiming to predict the final output sentences.
In the following, we elaborate the above four parts more in detail.

\subsection{Multi-Channel Sequence Processor}
The formats and semantics of the raw images and texts are usually rendered in different spaces.
To adapt them, we firstly input different modality sequences into a multi-channel sequential neural network.
Formally, for the image sequence $\{\bm{x}_{j}^I\}_{j=1}^{L}$, we feed it into a sequence model as follows:
\begin{eqnarray}\label{encoder}
\begin{aligned}
\bm{h}_{1}^I,\bm{h}_{2}^I,...,\bm{h}_{L}^I = g^I(\bm{x}_{1}^I,\bm{x}_{2}^I,...,\bm{x}_{L}^I),
\end{aligned}
\end{eqnarray}
where $g^I$ can be either recurrent neural network or transformer and we finally consider it by trading off the effectiveness and efficiency.
The output is a sequence of hidden embeddings $\bm{h}_{1}^I,\bm{h}_{2}^I,...,\bm{h}_{L}^I$.
Similarly, for the text sequence $\{\bm{x}_{j}^T\}_{j=1}^{L}$, we process it by a sequence model as follows:
\begin{eqnarray}\label{encoder}
\begin{aligned}
\bm{h}_{1}^T,\bm{h}_{2}^T,...,\bm{h}_{L}^T = g^T(\bm{x}_{1}^T,\bm{x}_{2}^T,...,\bm{x}_{L}^T),
\end{aligned}
\end{eqnarray}
\noindent where we implement $g^T$ with the same architecture as $g^I$, and their parameters are independently initialized and optimized in the learning process.
For highlighting the working flow of our idea, we delay the specifications of $g^I$ and $g^T$ in later sections.

\subsection{Spanning Influence Modeling}
Roughly speaking, our model is a sequence to sequence architecture.
However, unlike traditional tasks such as machine translation, where each input word usually corresponds to an output word rigorously, in our problem, the image or text may influence a span of the output sequence. {For example, in Figure~\ref{fig:intro}, image C influences two sentences, including "I could offer you a warm embrace" and "To make you feel my love".
}
In order to model such characters, we design a tailored module to capture the spanning influence of the input on the outputs.
Specifically, we let the hidden embedding derived in the above section influence the output sequences attentively.
Formally, for each $\bm{h}_{j}^I$, we separate it into $L$ parts, with each one corresponding to a sentence in $\bm{y} = \{\bm{y}_{k}\}_{k=1}^L$, that is:
\begin{eqnarray}\label{encoder}
\begin{aligned}
 \bm{h}_{j,k}^I = \alpha_{j,k}\bm{h}_{j}^I,
\end{aligned}
\end{eqnarray}
where $\alpha_{j,k}$ is an attention weight, and we implement it as:
\begin{eqnarray}\label{encoder2}
\begin{aligned}
\alpha_{j,k} = \frac{\exp{([\bm{W}\bm{h}_{j}^I]_k)}   }{\sum_{k'=1}^L \exp{([\bm{W}\bm{h}_{j}^I]_{k'})}  },
\end{aligned}
\end{eqnarray}
where $[\bm{a}]_k$ indicates the $k$-th element in vector $\bm{a}$, $\bm{W}$ is a weighting parameter projecting $\bm{h}_{j}^I$ into $L$-dimension.
Similarly, we derive the text partial hidden embeddings $\bm{h}_{j,k}^T~(k\in [1, L])$ for $\bm{h}_{j}^T$.

Above, we introduce flexibility in the correspondences between the inputs and outputs.
However, we argue that they should also follow some intuitive patterns. For example, if the distance between the input and output is large, then the influence should be small.
For example, although image C in Figure~\ref{fig:intro} influences the third and fourth sentences, the first sentence is not that relevant to it.
In order to encode such intuitions into our model, we further introduce a regularizer to constrain the attention weights.
Formally, we minimize the distance between $\bm{\alpha}_j = \{\alpha_{j,k}\}_{k=1}^L$ and a pre-defined distribution, which induces the following objective:
\begin{eqnarray}\label{dis-dis}
\begin{aligned}
L_{D} = D(\bm{\alpha}_j, \bm{\gamma}(j)),
\end{aligned}
\end{eqnarray}
where we implement $\bm{\gamma}(j)$ as the Gaussian distribution, and the mean and variance are set as $j$ and 1, respectively.
We empirically implement $D$  as the KL divergence.
However, it can be realized with other distribution distance functions to satisfy the specific scenarios.
By minimizing $\bm{\alpha}_j$ and $\bm{\gamma}(j)$, we regularize the attention weights with a prior, which encodes the intuition that a larger input-output distance should lead to a lower influence between them.

\begin{remark}
In this section, we design a tailored attention mechanism for our task, where the inputs and outputs correspond in an asynchronous manner, that is, each input may serve for several outputs. To achieve this goal, we use an attention mechanism to learn the soft correspondences between the inputs and outputs.
However, we do not let the attention weights blindly learn in the optimization process. Instead, we introduce prior knowledge on the distribution of the attention weights to narrow the exploration space, which can lead to a better converge rate and optimization solutions.
We believe the pre-defined distribution is not the ground truth, its function is to introduce prior knowledge to the attention learning process.
The advantages of the attention network itself such as the data adaption learning and expressiveness should not be ignored. Thus we use this prior distribution to guide the attention learning rather than using the distribution itself.
\end{remark}

\subsection{Multi-Modal Fusion Network}
Based on the above derived partial hidden embeddings $\bm{h}_{j,k}^I$ and $\bm{h}_{j,k}^T~(j\in [1, L], k\in [1, L])$, we fuse different modalities to derive the output of the encoder.
In specific, the output of the encoder is composed of $L$ embeddings, each of which encodes the topic, visual and textual information as follows:
\begin{eqnarray}\label{beta-dis-dis}
\begin{aligned}
\bm{e}_{k} = \sum_{j=1}^L\sum_{j'=1}^L g^F(\bm{t},\bm{h}_{j,k}^I,\bm{h}_{j',k}^T),~\quad k\in [1, L],
\end{aligned}
\end{eqnarray}
where $\bm{e}_{k}$ is computed by iterating the influence of hidden embeddings from different steps on the $k$-th steps.
For each pair of steps $(j,j')$, different modalities are combined in an attentive manner:
\begin{eqnarray}\label{beta1}
\begin{aligned}
g^F(\bm{t},\bm{h}_{j,k}^I,\bm{h}_{j',k}^T) = \beta^{t}_k\bm{t} + \beta^{I}_k\bm{h}_{j,k}^I + \beta^{T}_k\bm{h}_{j',k}^T,
\end{aligned}
\end{eqnarray}
where $\beta^{t}_k$, $\beta^{I}_k$ and $\beta^{T}_k$ are the attention weights computed as:
\begin{eqnarray}\label{beta2}
\begin{aligned}
&\beta^{i}_k = \frac{\exp(\bm{W^t}\bm{s}_i)}{\sum_{i'\in \{t, I, T\}} \exp(\bm{W^{i'}}\bm{s}_{i'})}, \\
\end{aligned}
\end{eqnarray}
where $\bm{s}_{t} = \bm{t},~~~\bm{s}_{I} = \bm{h}_{j,k}^I,~~~\bm{s}_{T} = \bm{h}_{j,k}^T$.
\begin{remark}
(\romannumeral1) Intuitively, for the same output sentence, different modalities may play different roles. As a result, we deploy an attention mechanism when combining them in equation~(\ref{beta1}). 
(\romannumeral2) If one compares the attentions used in equation~(\ref{encoder}) and~(\ref{beta1}), she may find that the former is deployed across different steps within the same modality, while the latter aims to capture the contributions of different modalities at the same step. Such design actually forms a 2D attention mechanism, which is expected to model the influences of different positions and modalities in a finer-grained manner.
\end{remark}

\subsection{Experience Enhanced Sentence Decoder}
In the above sections, we have detailed the working principles of the encoder.
In this section, we describe how to generate the outputs based on the embeddings $\bm{e}_{k}~(k\in [1, L])$.
Straightforwardly, one can merge different $\bm{e}_{k}$'s, and use the result as the prompt~\cite{brown2020language,lester2021power} to directly induce all the output sentences. 
However, such a strategy can be sub-optimal for preserving the sequential semantics of the input, since the ordered information can be weakened by the merging operation.
To solve the above problem, we let each experience embedding $\bm{e}_{k}$ influence the output sentence separately.
Formally, we add $\bm{e}_{k}$ with the word embedding at each step, that is:
\begin{eqnarray}\label{decod}
\begin{aligned}
&\bm{\pi}_{k,i+1} = \text{DEC}(\bm{\pi}_{<k,<i}, \bm{e}_{k}+\bm{w}_{k,i}),~i\in[1,d_{k}],
\end{aligned}
\end{eqnarray}
where we denote the decoder by DEC.
$\bm{w}_{k,i}$ is the word embedding of the $i$-th word in the $k$-th sentence. 
$\bm{\pi}_{k,i}$ is the $i$-th token in the $k$-th sentence, where $\bm{\pi}_{<k,<i}$ denotes all the tokens before $\bm{\pi}_{k,i}$.
$d_{k}$ is the length of the $k$-th output sentence.

\subsection{Model Optimization}
Massive previous works suggest that introducing negative samples leads to better optimization results~\cite{mao2016generation,lee2020contrastive}.
Following the same strategy, we maximize the probability of generating the ground truths from the positive inputs and simultaneously minimize that of producing the ground truths from the negative inputs.
Given the dataset $\mathcal{S} = \{(\bm{t}_i, \{(\bm{x}_{i,j}^I, \bm{x}_{i,j}^T)\}_{j=1}^{L}), \bm{y}_i\}_{i=1}^N$, the learning objective is:
\begin{eqnarray}\label{lc}
\begin{aligned}
\mathcal{L} = \sum_{i=1}^N \{\log \sigma(f(\bm{t}_i, \bm{x}_{i}, \bm{y}_i)) + \sum_{\bm{x}_{i}^-\in \bm{O}_i^-} \log \sigma(1-f(\bm{t}_i, \bm{x}_{i}^-, \bm{y}_i))\},
\end{aligned}
\end{eqnarray}
where 
$\bm{x}_{i} = \{(\bm{x}_{i,j}^I, \bm{x}_{i,j}^T)\}_{j=1}^{L}$ is the input sequence of the image-text pairs.
$\bm{O}_i^-$ is the set of negative inputs for $\bm{x}_{i}$.
$f(\bm{t}_i, \bm{x}_{i}, \bm{y}_i)$ is the probability of generating $\bm{y}_i$ based on $\bm{t}_i$ and $\bm{x}_{i}$.

\noindent\textbf{Curriculum Negative Sampling.} Previous negative sampling strategies are mostly designed for a single input.
In our task, the input is a sequence, which brings additional difficulties to conduct negative sampling.
As the sequence becomes longer, the negative sample space is exponentially enlarged, making it impossible to select all the negative samples. 
To better learn our model, we select the negative samples in a curriculum learning manner.
Our general idea is to first learn the most negative samples for better initializing the model optimization. Once the model has learned enough patterns to handle the most negative ones, we gradually introduce harder samples near the positive and negative boundaries.
More specifically, we construct samples with 5 levels with the relevance rank of the input image/text with the output. 
Level-5 means the most relevant input, and level-1 indicates the input and output are the most irrelevant.
In the training process, we first train the model with Level-5 and Level-1 instances, and then incorporate Level-4 and Level-2 into the positive sample set and $\bm{O}_i^-$, respectively. Level-3 ones are taken as negative samples at last.

\subsection{Detailed Implementation of $g^T$ and $g^I$}
Now we present the implementation of $g^T$ and $g^I$.
In specific, we first use the multi-modal pre-trained model WenLan~\cite{huo2021WenLan} (similar to OpenAI CLIP but trained on Chinese data) to project the image and text separately into the same semantic space:
\begin{align}
\begin{split}
    \bm{v}^{t}, \bm{v}_{j}^{I}, \bm{v}_{j}^{T} &= \textsc{WenLan}(t), \textsc{WenLan}(x_{j}^{I}), \textsc{WenLan}(x_{j}^{T}).
\end{split}
\end{align}
Meanwhile, we adopt a linear layer, followed by a layer normalization $\textsc{Ln}(\cdot)$, to project the topic word into an embedding as follows:
\begin{align}
    \hat{\bm{v}}^{t} = \textsc{Ln}\left(\textrm{Linear}(\bm{v}^{t})\right),
\end{align}

To capture the sequential information of the experiences, we process the embedded image and text with the Gated Recurrent Units (GRU)~\cite{chung2014empirical}.
The input image sequence $\mathbf{V}^{I}=\{\bm{v}_{j}^{I}|j=1,\ldots,L\}$ and text sequence $\mathbf{V}^{T}=\{\bm{v}_{j}^{T}|j=1,\ldots,L\}$ are encoded as:
\begin{align}
\begin{split}
    \hat{\mathbf{V}}^{I}, \hat{\mathbf{V}}^{T} &= \textsc{Ln}\left(\textrm{GRU}_\textsc{Text}(\mathbf{V}^{T})\right), \textsc{Ln}\left(\textrm{GRU}_\textsc{Img}(\mathbf{V}^{I})\right),
\end{split}
\end{align}

To lower the risk of over-fitting and flatten the feature distribution, we deploy normalization upon the outputs.
The normalized embeddings are then fed into spanning influence modeling layers.

\section{Related Work}
In this section, we review the previous work from two perspectives. 
For the task, our paper is highly relevant to the problems of lyrics and poetry writing, as well as multi-modal generation.
For the technique, our method is built upon the prosperity of recent multi-modal representation models.
In the following, we introduce the classical models and recent advances in these fields more in detail.

\subsection{Lyrics and Poetry Generation}
Lyrics generation and poetry writing are two typical AI creation tasks, where the generated texts need to follow some formats~\cite{li2020songnet} and rhymes~\cite{xue2021deeprapper}.
Early works on lyrics generation are mostly based on constraints~\cite{barbieri2012markov,addanki2013unsupervised} or retrieval-based methods, attempting to generate by matching the best relevant rear lines with the prior ones~\cite{malmi2016dopelearning}. 
Later studies use neural networks like Long Short-Term Memory (LSTM)~\cite{potash2015ghostwriter,watanabe2018melody} or autoencoder~\cite{liang2018attae,nikolov2020rapformer} to handle this task, adding hierarchical attention mechanism in the  decoders~\cite{wang2019theme,fan2019hierarchical}. Recently, pre-trained language models can provide better conditional results~\cite{zhang2020youling} and considering more rhymes and rhythms~\cite{xue2021deeprapper}. 
However, none of these efforts take images as inputs or conditions. 

In the task of poetry generation, early models mainly focus on the keywords expansion and modeling the poet's intents~\cite{wang2016chinese,yi2018chinese}, until when evolved to a milestone with the advent of the giant pre-trained language model like GPT~\cite{liao2019gpt,zou2021controllable}.
Besides text information, other attempts lie in image-inspired poetry generation. 
These works~\cite{cheng2018image,xu2018images} employ visual input to simulate the scenery perception process of humans. 
Despite many promising results, most of them identify keywords, e.g., objects or sentiments, from an image and adopt the keywords as input to influence the poem generation process.
Basically, these methods generated poems from a single image input, which significantly differs from our model that tries to capture the sequential semantics from a series of images. 
\citet{liu2018images2poem} proposed Images2Poem to generate classical Chinese poetry from image streams by selecting representative images from a stream and adopting an adaptive self-attention mechanism to decode, which is similar to our work but the constructed images (about 20 images per poem) are mainly objects mentioned in a poem. 
Different from them, we aim to generate text with a few images and aligned texts as a simulation of human embodied experiences because not all experiences, such as feelings, can be well visualized. We summarize the differences between ours and the most related works in Table~\ref{Comparison}.

\subsection{Multi-modal Generation}

As a typical task of multi-modal generation, multi-modal summarization~\cite{wang2016low,zhu2018msmo,dai2022enabling} generates text summary by adopting multi-modal data. However, the generated summary is highly depended on source text, which is different from our topic-aware generation task. 
Another task related to ours is visual storytelling, which takes multiple sequential images as input and aims to generate coherent stories~\cite{huang2016visual,lukin2018pipeline}.
To solve this problem, many works leverage CNNs to encode image streams and RNN-liked blocks to generate story sentences~\cite{huang2016visual,yu2017hierarchically,li2019informative}, or with hierarchical structures~\cite{wang2019hierarchical,su2021bert,fan2021visual} accompanied by some dedicated designs on the attention mechanisms~\cite{braude2021towards}.
Although some works~\cite{li2020topic} endowed the model the ability to adapt to topic or incorporated videos for visual storytelling~\cite{li2019video}, few works studied using both topic and paired image-text input like our setting, which is a more realistic simulation of experiences.

\begin{table}[t]
\centering
  \caption{Comparison of different poetry generation methods. A check mark indicates the information of this kind of input can be processed (Tp.: Topic, Img.: Image, Mul-Img: Multiple Images, Ex-T.: Extra Text, MM.: Multi-modal Modeling).}
  \label{Comparison}
\begin{tabular}{lccccc}
    \toprule
    \textbf{Models} & \textbf{Tp.} & \textbf{Img.} & \textbf{Mul-Img.} & \textbf{Ex-T.} & \textbf{MM.} \\ \hline
    \textbf{Plan}~\cite{wang2016chinese} & \Checkmark &  &  & \Checkmark & \\
    \textbf{WM}~\cite{yi2018chinese} & \Checkmark &  &  &  &  \\
    \textbf{GPT-2}~\cite{radford2019language} & \Checkmark &  &  & \Checkmark &  \\
    \textbf{iPrompt}~\cite{radford2019language} & \Checkmark &  &  & \Checkmark &  \\
    \textbf{I2P-GAN}~\cite{liu2018beyond} &  & \Checkmark &  &  & \Checkmark \\
    \textbf{Images2Poem}~\cite{liu2018images2poem} &  & \Checkmark & \Checkmark &  & \Checkmark \\
    \textbf{MMTG (ours)} & \Checkmark & \Checkmark & \Checkmark & \Checkmark & \Checkmark \\
    \bottomrule
\end{tabular}
\end{table}

\subsection{Multi-modal Representation and Learning}
With the ever prospering of the web technologies, the internet has accumulated a large amount of multi-modal information. 
To take advantage of this, researchers have designed a lot of promising methods to learn the representations of different modalities.
In~\cite{vo2019composing}, a new approach named Text Image Residual Gating is designed, aiming to combine both image and text for image retrieval. 
Later, pre-trained methods have shined in many multi-modal tasks owing to their great capacity to learn representations from vision and language inputs~\cite{lu2019vilbert, li2020unicoder, chen2020uniter,li2020unimo}. 
OpenAI CLIP~\cite{radford2021learning} is trained upon a dataset of 0.3 billion image-text pairs with multi-task objectives, adopting contrastive learning to bridge the visual-language gap.
A similar multi-modal pre-trained model named WenLan~\cite{huo2021WenLan} is a two-tower one within the cross-modal contrastive learning framework. It is trained on 0.65 billion image-text pairs and can encode image and text separately into the same semantic space. We adopt this as one of our initiations since WenLan released a Chinese model.

\section{Experiments}
\label{sec:experiments}
In this section, we conduct extensive experiments to verify the effectiveness of our model. We first elaborate on how we prepare a dataset and then introduce implementation details and evaluation metrics.
At last, we present and analyze experimental results.

\subsection{Data Preparation}
\label{data_preparation}
To the best of our knowledge, there is no publicly available dataset for our task. 
In order to evaluate our model, and demonstrate its effectiveness, we manually labeled a new dataset.

In specific, our labeling process includes three phases:
(1) Crawling the output texts.
In our dataset, the output texts are crawled from a famous Chinese song website~\footnote{\url{https://music.163.com}}.
For each ten-line passage of a song, we separate it into five sentences, each of which corresponds to a $\bm{y}_{i,k}$ in $\bm{y}_i = \{\bm{y}_{i,k}\}_{k=1}^L$ (defined in Section~\ref{formu}), that is, in our dataset, $L=5$. The title of the song is regarded as the topic $\bm{t}_i$.
(2) Collecting the input image-text pairs.
For each output sentence $\bm{y}_{i,k}$, we collect the input image-text pairs $(\bm{x}_{i,k}^I, \bm{x}_{i,k}^T)$ from a dataset called GraphMovie~\cite{chen2019neural}, where the movie screenshots and story-telling text are regarded as the image ($\bm{x}_{i,k}^I$) and text ($\bm{x}_{i,k}^T$) information, respectively. 
For each $\bm{y}_{i,k}$, we retrieve 3 image-text pair candidates.
(3) By the above two steps, we can already build the dataset defined in Section~\ref{formu}, that is, $\mathcal{S} = \{(\bm{t}_i$, $\{(\bm{x}_{i,j}^I, \bm{x}_{i,j}^T)\}_{j=1}^{L}), \bm{y}_i\}_{i=1}^N$.
To facilitate our negative sampling strategy, we label the relevance between the image-text pairs and the corresponding output sentences. 
We further train a \textit{WenLan-ranker} on the labeled data to re-rank the 3 candidates, so that to construct 5 different levels of training samples. Suppose the most relevant image-text pair is denoted by Rank1, and the most irrelevant one is denoted by Rank3, then we define our 5-level samples as follows:
\begin{itemize}[leftmargin=*]
    \item Level-5 (most positive): it contains 5 Rank1 image-text pairs (that is, for each step, we select the most relevant image-text pair). 
    \item Level-4: it contains 3 Rank1 image-text pairs, 1 Rank3 image-text pair, and 1 negative image-text pair randomly sampled from the unlabeled samples; 
    \item Level-3: it contains 5 Rank3 image-text pairs; 
    \item Level-2: it contains 1 Rank1, 1 Rank3, and 3 negative samples; 
    \item Level-1 (most negative): it contains 5 negative samples. 
\end{itemize}

The unit of our built dataset is $(\bm{t}_i, ((\bm{x}_{i,j}^I, \bm{x}_{i,j}^T)\}_{j=1}^{L}), \bm{y}_i)$, which is called as an \textit{e-passage}.
In our dataset, we have 46,192 e-passages in total and finally construct 220,960 (= 44,192 × 5) samples for curriculum learning. 
To verify whether the automatic evaluation results are consistent with human ratings, we use 50 of them as the test set~\footnote{This is owing to the cost of human evaluation. We have also made evaluations on a test set containing 2,960 samples on the automatic metrics. The conclusions are almost the same as the original test set.}, and the others are left for training.
The detailed statistic of our dataset can be seen in Table~\ref{tab:dataset}.

\begin{table}[t]
  \centering
  \caption{Statistics of the lyrics corpus and e-passage dataset.}
  \label{tab:dataset}
  \begin{tabular}{lr}
  \toprule
  overall\\
  \quad\# of lyrics corpus & 410,335 \\
  \hline
  simulating passages with experiences\\
  \quad\# of e-passages & 46,192 \\
  \hline
  training WenLan-ranker\\
  \quad\# of e-passage manually labeled & 1,950 \\
  \hline
  automatically labeled for overall training \\
  \quad\# of e-passages & 44,192 \\
  \hline
  a test set for evaluation\\
  \quad\# of e-passages & 50\\
  
  \bottomrule
  \end{tabular}
\end{table}

\subsection{Training Details}
\label{sec:Setup}
We initialize all the inputs with the  WenLan~\footnote{We adopt the original model of WenLan: \url{https://github.com/BAAI-WuDao/BriVL}.} embeddings, which are of 2,048 dimensions. 
When fine-tuning the \textit{WenLan-ranker}, we adopt the candidates with the highest ratings as positive samples and the other candidates as negative samples. To avoid overfitting, WenLan-ranker is fine-tuned for 1 epoch with a learning rate of 1e-5 and batch size of 32.

In the pre-training phase, the GPT-2 model~\footnote{\url{https://github.com/Morizeyao/GPT2-Chinese/tree/master}} is initialized with the parameters pre-trained on the Chinese Clue Corpus~\footnote{\url{https://huggingface.co/uer/gpt2-chinese-cluecorpussmall}}. 
It contains 12 layers with 12 attention heads, and the word embedding dimension is set as 768. The vocabulary size is 13,317 and all tokens in the vocabulary are encoded by WenLan. The  2,048-dimension inputs are fed into the projector and mapped to 768-dimension embeddings for GPT-2, and the hidden state dimension is 512. The projector and GPT-2 are trained on our pre-training lyrics corpus for 1 epoch, while the learning rate is 5e-5 and batch size is 32.

We adopt 1-layer GRU and set the hidden sizes of the multi-channel sequence processor and multi-modal fusion network to 512. Self-attention heads of spanning influence modeling is set to 4. The model is trained using a learning rate of 1e-5 with batch size 96 for 5 epochs.
During decoding, we apply top-k and top-p sampling to generate texts by setting k to 10 and p to 0.7 at a temperature of 1.1. The repetition penalty is set to 1.5 to reduce repeating.

\begin{table}[t]
\caption{Results of automatic evaluation on two baselines and our model. The metrics B.-2, Dist.-2 and B.S. stand for BLEU-2, Distinct-2 and BERTScore. $\dagger$ and $\star$ denote significant improvements over the baseline results with $p$-value$<0.01$ and $p$-value$<0.05$ in t-test, the same below.}
  \centering
  \label{table:autoresults}
    \setlength{\tabcolsep}{1.3mm}{
    \begin{tabular}{@{}lccccc@{}}
    \toprule
    \textbf{Methods} & \textbf{B.-2} & \textbf{Dist.-2} & \textbf{B.S.} & \textbf{NNR-1} & \textbf{NNR-2} \\ \hline
    \textbf{GPT-2}~\cite{radford2019language} & 0.075 & 0.583$^{\dagger}$ & 0.576$^{\dagger}$ & - & -\\
    \textbf{Images2Poem}~\cite{liu2018images2poem} & 0.066$^{\dagger}$ & 0.660$^{\dagger}$ & 0.585$^{\dagger}$ & 0.023$^{\dagger}$ & 0.041$^{\dagger}$ \\
    \textbf{MMTG (ours)} & \textbf{0.076} & \textbf{0.743} & \textbf{0.595} & \textbf{0.315} & \textbf{0.411} \\ 
    \bottomrule
    \end{tabular}}
\end{table}

\begin{table}[t]
\caption{Results of human evaluation. Due to the subjectivity, we report the average scores of these metrics.}
  \centering
  \label{table:results}
    \setlength{\tabcolsep}{0.5mm}{
    \begin{tabular}{@{}lcccc@{}}
    \toprule
    \textbf{Methods} & \textbf{Relevance} & \textbf{Coherence} & \textbf{Meaning} & \textbf{Overall} \\ \hline
    \textbf{Plan}~\cite{wang2016chinese} & 1.62$^{\dagger}$ & 2.27$^{\dagger}$  & {2.27}$^{\dagger}$ & 2.11$^{\dagger}$ \\
    \textbf{WM}~\cite{yi2018chinese} & 1.79$^{\dagger}$ & 2.33$^{\dagger}$ & {2.32}$^{\dagger}$ & 2.16$^{\dagger}$ \\
    \textbf{GPT-2}~\cite{radford2019language} & 1.46$^{\dagger}$ & 1.67$^{\dagger}$ & 1.88$^{\dagger}$ & 1.60$^{\dagger}$ \\
    \textbf{iPrompt}~\cite{zou2021controllable} & {1.97} & {2.42}$^{\dagger}$ & {2.45}$^{\star}$ & {2.34} \\
    \textbf{I2P-GAN}~\cite{liu2018beyond} & 1.40$^{\dagger}$ & 1.80$^{\dagger}$ & 2.02$^{\dagger}$ & 1.65$^{\dagger}$ \\
    \textbf{Images2Poem}~\cite{liu2018images2poem} & 1.52$^{\dagger}$ & 1.92$^{\dagger}$ & 2.14$^{\dagger}$ & 1.83$^{\dagger}$ \\ \hline
    \textbf{MMTG (ours)} & \textbf{2.11} & \textbf{2.68} & \textbf{2.57} & \textbf{2.47} \\ \bottomrule
    \end{tabular}}
\end{table}

\subsection{Baselines}
We compare our model with the following representative methods:
\begin{itemize}[leftmargin=*]
    \item \textbf{Plan}~\cite{wang2016chinese}: a planning-based poetry generation method.
    \item \textbf{WM}~\cite{yi2018chinese}: a model based on a working memory mechanism that dynamically generates poetry with coherence guarantees.
    \item \textbf{GPT-2}~\cite{radford2019language}: an auto-regressive language model that generates texts based on a given prompt. 
    \item \textbf{iPrompt}~\cite{zou2021controllable}: a recently proposed state-of-the-art method that predicts the prompt during beam search for better controlling the text generation. It uses 302GB of general text data for training a base language model having 2.86 billion parameters. The base model has the GPT framework with its transformer model substituted to Transformer-XL.
    \item \textbf{I2P-GAN}~\cite{liu2018beyond}: an adversarial reinforcement learning model generating poetry based on an image. To fit our setting, we generate two sentences for each image and translate English into Chinese. 
    \item \textbf{Images2Poem}~\cite{liu2018images2poem}: a seq2seq model that generates poems taking image streams as inputs. It is based on LSTM without pre-training, so it is hard to converge on the data of diverse lyrics of various lengths. Thus we implement their work with our codes based on attention blocks but inputting only images and outputting corresponding lyrics with the decoder pre-trained.
\end{itemize}

\begin{figure}[t]
    \centering
    \includegraphics[width=1.\linewidth]{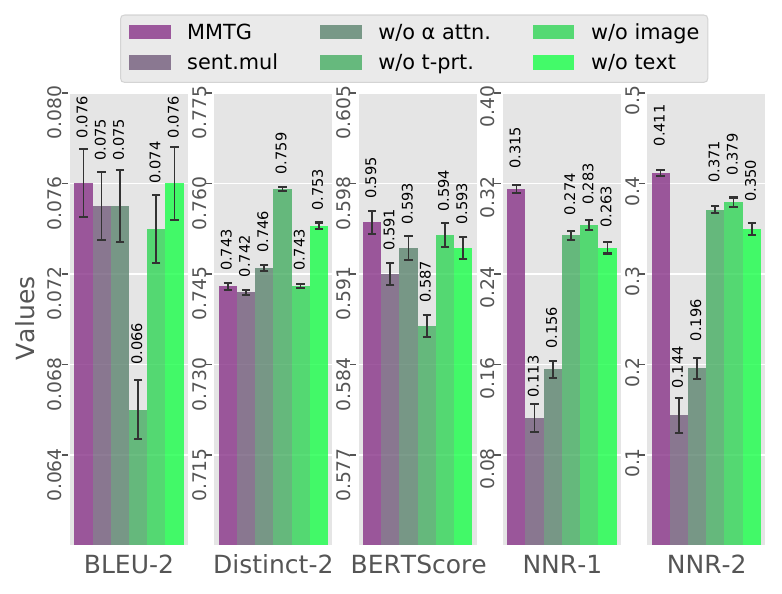}
    \caption{Results of ablation study on different variants. ``$\alpha$ attn.'' and ``t-prt.'' refer to $\alpha$-attention and t-prompt.}
    \label{fig:abaltion_var}
\end{figure}

\subsection{Evaluation Metrics}
Different from other generation tasks like machine translation, in AI creation, human-centered metrics are more important. Thus we adopt both automatic and human metrics for evaluation.

\noindent \textbf{Automatic Metrics.} We adopt the following widely used metrics:
\begin{itemize}[leftmargin=*]
    \item \textbf{BLEU: } Bilingual Evaluation Understudy (BLEU)~\cite{papineni2002bleu} which determines the similarity of two sentences based on n-grams.
    \item \textbf{Distinct: } we adopt Distinct~\cite{li2015diversity} considering the ratio of unique n-grams to evaluate the diversity among the generated texts.
    \item \textbf{BERTScore: } n-gram models may fail to capture distant dependencies and semantic ordering change, thus BERTScore~\cite{bert-score} has recently been proposed and widely used. It leverages the contextual embeddings from BERT and matches words in candidate and reference sentences by cosine similarity.
    \item \textbf{NNR: } to measure the differences of output texts with different orders of the input image-text sequence, we propose calculating the New N-grams Rate (NNR) as follows:
    \begin{align}
        NNR = \frac{uniq(Y)-uniq(X)}{uniq(X)\cup uniq(Y)}
    \end{align}
    where $uniq(\cdot)$ is the number of unique n-grams of a set, and $Y$, $X$ here stand for output texts with disordered experience input and ordered experience input respectively. The larger the NNR, the more sensitive is the algorithm to the order of input sequence.
\end{itemize}
To reduce randomness and get reliable results, all these metrics are the average values by generating 10 samples for each test data.

\noindent \textbf{Human Rating Criteria.} Following previous work~\cite{yi2018chinese, liu2018images2poem, shen2020compose}, some criteria for human evaluation are applied as follows.
\begin{itemize}[leftmargin=*]
    \item \textbf{Relevance:} how a generated text is relevant to a given topic.
    \item \textbf{Coherence:} whether an output is coherent across lines and semantically fluent through the whole passage.
    \item \textbf{Meaning:} whether a generated text has a certain meaning, including clear informative delivery.
    \item \textbf{Overall:} overall quality of a generated text.
\end{itemize}

Each criterion is judged on a 5-point scalar ranging from 1 (worst) to 5 (best). Three annotators are asked to rate all results independently. The texts generated by different methods of each test data are shuffled to remove position bias, displayed on the same page to obtain consistent relative judgments and hidden method names for fairness. The 3 judge ratings are averaged as the final rating.

\subsection{Results and Analysis}

\subsubsection{Comparison with Baselines}
We compare our MMTG model with GPT-2 and Images2Poem in automatic evaluation on our dataset. Results are shown in Table~\ref{table:autoresults}. Our MMTG outperforms the two baselines in BLEU-2, BERTScore, and Distinct-2. Improvement over GPT-2 indicates fusing multi-modal experiences brings additional benefits than training a text-only language model. This verifies our assumption that writers may involve multi-modal experiences in creation. Improvement over Images2Poem indicates both visual and textual information helps. Furthermore, in terms of NNRs, our proposed model can generate different lyrics with the order change of input images while Images2Poem is not sensitive. It indicates MMTG achieves using experiences in a sequential way.

We compare our MMTG model with these baselines: four (i.e., Plan, WM, GPT-2, and iPrompt) take test topics as input and two (I2P-GAN and Images2Poem) take image experiences as input. 
Human evaluation results showed in Table~\ref{table:results} indicates that our model performs the best in terms of all metrics. The improvements over Plan, WM, GPT-2, I2P-GAN, and Images2Poem in all criteria are also statistically significant. We find that it is not trivial to take the advantage of multi-modal input because actually the other two methods with images input, i.e., I2P-GAN and Images2Poem, work worse than iPrompt, Plan, and WM methods with text input only. iPrompt benefits from a base GPT-3 like language model, which has 2.86 billion parameters and was pre-trained on 302GB of general text data. It performs well in \textit{Relevance} and \textit{Overall}, while our MMTG model still has improvements but the difference is not statistically significant. This indicates again the difficulty that we use a small amount of multi-modal information for training to beat a generation model with text input only but based on a large-scale foundation model. Our proposed method works significantly better than iPrompt in terms of \textit{Coherence} and \textit{Meaning}. This may be because image-text pairs provide richer details to enhance \textit{Meaning} and the model with sequential design helps improve \textit{Coherence}.

\begin{figure*}[t]
    \centering
    \includegraphics[width=.98\linewidth]{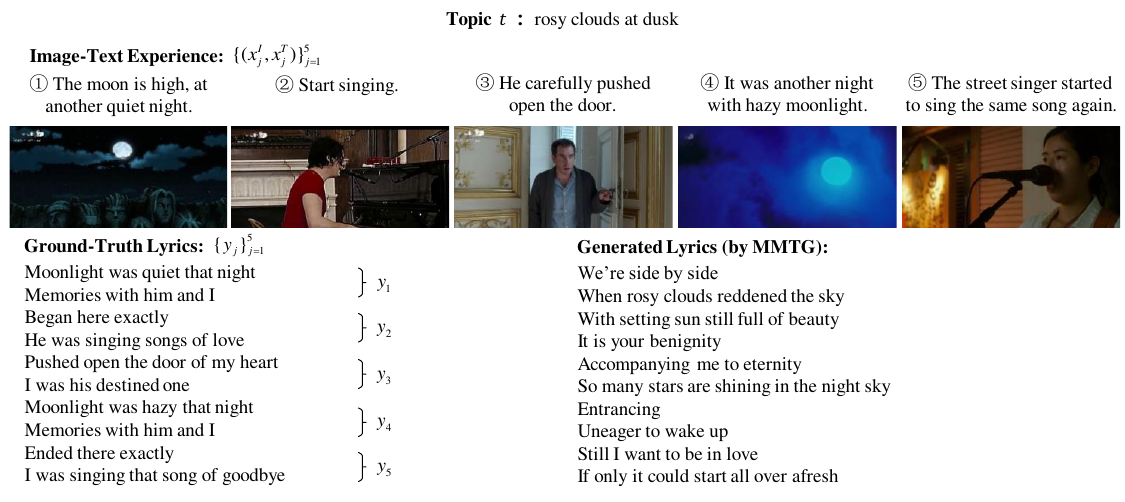}
    \caption{A case of a topic, five image-text pairs as experiences, ground-truth, and the generated lyrics by our MMTG model. }
    \label{fig:case}
\end{figure*}

A case is shown in Figure~\ref{fig:case}. It demonstrate the good ability of MMTG to model multi-modal information. 
For example, there are corresponding expressions like ``rosy clouds reddened the sky'', `` setting sun'', `` stars'', ``entrancing'' in the results by integrating multimodal information like ``dusk'', ``moon'', ``open the door'' in the input. 
Besides, the generated text ``start all over afresh'' and ``Ended there exactly'' in the ground-truth amazingly express the similar idea in different words.
This demonstrates the good generation ability of our proposed MMTG model.

\begin{figure}[t]
    \centering
    \includegraphics[width=.99\linewidth]{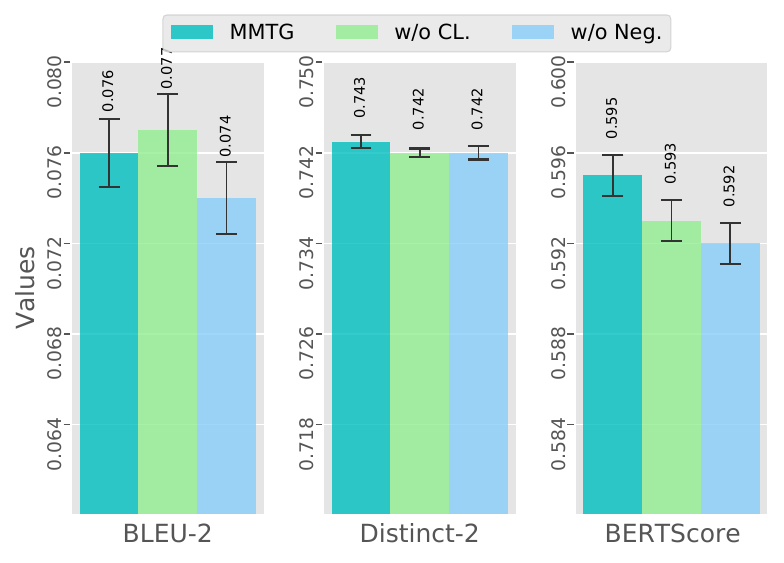}
    \caption{Results of ablation study on different training strategies. ``CL.'' and ``Neg.'' are short for curriculum learning and negative samples respectively.}
    \label{fig:abaltion_cl}
\end{figure}

\subsubsection{Ablation Study of Different Structures}
To compare different structures, we compare our model with its five variants, that is:
\begin{itemize}[leftmargin=*]
    \item \textbf{MMTG w/ sent.mul}: a variant that word embeddings and experience embeddings are multiplied at sentence level in equation~\ref{decod}.
    \item \textbf{MMTG w/o $\alpha$-attention}: a variant without the pre-defined distribution in equation~\ref{dis-dis}.
    \item \textbf{MMTG w/o t-prompt}: a variant without topic prompt.
    \item \textbf{MMTG w/o image}: a variant without image experience inputs.
    \item \textbf{MMTG w/o text}: a variant without text experience inputs.

\end{itemize}

The results are presented in Figure~\ref{fig:abaltion_var}. Overall, our MMTG performs the best in terms of all metrics except for Distinct-2. This indicates our proposed ideas do contribute to better quality and sensitiveness to the input order. In specific, we observe the following different influences: 1) using multiplying experience embedding to word embedding in the decoder part works as not as good as adding in terms of BERTScore and NNRs. This indicates adding is a better choice to make experience embedding influence generation content and ordering; 2) the big drop without $\alpha$-attention in terms of NNRs indicates the mechanism can indeed capture sequential information. 3) removing the topic prompt dramatically decreases BLEU-2 and BERTScore, which represents without the constraint of a topic the generated lyrics may drift to diverse but irrelevant topics. 4) both image and text experiences are useful in creating better lyrics in terms of BLEU-2, BERTScore, and NNRs. This proves visual and textual experiences have complementary information.

\subsubsection{Ablation Study of Different Optimization Strategies}
We train our model with different optimization strategies for comparison:
\begin{itemize}[leftmargin=*]
    \item \textbf{MMTG w/o CL}: a model without curriculum learning.
    \item \textbf{MMTG w/o Neg}: a model without negative sampling. Here only Level-5 samples are used.
\end{itemize}

We present the results in Figure~\ref{fig:abaltion_cl}. Our model MMTG performs the best on Distinct-2 and BERTScore. This indicates that both negative sampling and curriculum learning have their positive contributions. Adding negative samples somehow makes the model "less sure" and generates more diverse results, while still expressing accurate meanings. Without curriculum learning, both Distinct-2 and BERTScore drop, and it is comparable on BLEU-2. It indicates that learning in an ``easy-to-hard'' manner can make the model better distinguish positive samples from negative ones.

\section{Conclusion and Future Work}
Multi-modal text generation is receiving increasing attention, but a simulation of human creation of literary works processing multi-modal information for topic-aware literary text generation has not been well studied. In this work, we propose a multi-modal seq2seq architecture named MMTG to solve this issue.
We model visual and textual input as experiences and use these interacted experience embeddings to correspond with different sentences under the constraint of topic for an auto-regressive generation. 
We design a novel curriculum negative sampling method to learn the parameters in an ``easy to hard'' manner. 
Experimental results on both automatic and human evaluation indicate the effectiveness of MMTG. 
Detailed analysis indicates experience embeddings and curriculum negative sample learning contribute the most to our proposed model. 

This paper actually advances toward more realistic human creation processes.
However, there is still much room left for improvement.
In the future, we plan to integrate the processes of experiences retrieval and text generation, where the output may provide valuable supervision signals to better guide the experience selection.

\begin{acks}
This work was supported by Beijing Outstanding Young Scientist Program NO. BJJWZYJH012019100020098, Beijing Key Laboratory of Big Data Management and Analysis Methods, Intelligent Social Governance Platform, the Research Seed Funds of School of Interdisciplinary Studies, Major Innovation \& Planning Interdisciplinary Platform for the “Double-First Class” Initiative of Renmin University of China. We acknowledge the anonymous reviewers for their helpful comments and also thank Ershan Wang for helping us translate our cases. Ruihua Song is the corresponding author.
\end{acks}

\bibliographystyle{ACM-Reference-Format}
\bibliography{reference}

\appendix

\end{document}